\documentclass[11pt]{article}

\usepackage[final]{acl}

\usepackage{times}
\usepackage{latexsym}

\usepackage[T1]{fontenc}
\usepackage[utf8]{inputenc}

\usepackage{microtype}

\usepackage{inconsolata}

\usepackage{graphicx}
\usepackage{multirow}

\usepackage{algorithm}
\usepackage{algorithmic}
\usepackage{amsmath}
\usepackage[most]{tcolorbox}
\newtcolorbox{promptbox}{
  colback=gray!5,
  colframe=black!60,
  boxrule=0.5pt,
  arc=2pt,
  left=6pt,
  right=6pt,
  top=6pt,
  bottom=6pt
}
\usepackage{makecell}
\usepackage{booktabs}
\usepackage{makecell}

\usepackage{booktabs}
\usepackage{makecell}
\usepackage{adjustbox}

\title{From Human Cognition to Neural Activations: Probing the Computational Primitives of Spatial Reasoning in LLMs}

\author{
 \textbf{Jiyuan An\textsuperscript{1,2}},
 \textbf{Liner Yang\textsuperscript{1,2,*}},
 \textbf{Mengyan Wang\textsuperscript{1,2}},
 \textbf{Luming Lu\textsuperscript{3}},
\\
 \textbf{Weihua An\textsuperscript{1}},
 \textbf{Erhong Yang\textsuperscript{1,2}},
\\
\\
 \textsuperscript{1}National Language Resources Monitoring and Research Center for Print Media, Beijing Lang-\\uage and Culture University, Beijing, China
 \\
 \textsuperscript{2}School of Information Science, Beijing Language and Culture University, Beijing, China
 \\
 \textsuperscript{3}School of artificial intelligence, Beijing normal university, Beijing, China
\\
}

\begin{document}
\maketitle

%%% 摘要
\begin{abstract}
As spatial intelligence becomes an increasingly important capability for foundation models, it remains unclear whether large language models’ (LLMs) performance on spatial reasoning benchmarks reflects structured internal spatial representations or reliance on linguistic heuristics. We address this question from a mechanistic perspective by examining how spatial information is internally represented and used. Drawing on computational theories of human spatial cognition, we decompose spatial reasoning into three primitives, relational composition, representational transformation, and stateful spatial updating, and design controlled task families for each. We evaluate multilingual LLMs in English, Chinese, and Arabic under single pass inference, and analyze internal representations using linear probing, sparse autoencoder based feature analysis, and causal interventions. We find that task relevant spatial information is encoded in intermediate layers and can causally influence behavior, but these representations are transient, fragmented across task families, and weakly integrated into final predictions. Cross linguistic analysis further reveals mechanistic degeneracy, where similar behavioral performance arises from distinct internal pathways. Overall, our results suggest that current LLMs exhibit limited and context dependent spatial representations rather than robust, general purpose spatial reasoning, highlighting the need for mechanistic evaluation beyond benchmark accuracy.
\footnote{Our code is available at <It will be published after the paper is accepted.>}
\end{abstract}

%%% 1
\section{Introduction}

\begin{figure}[t]
  \includegraphics[width=\columnwidth]{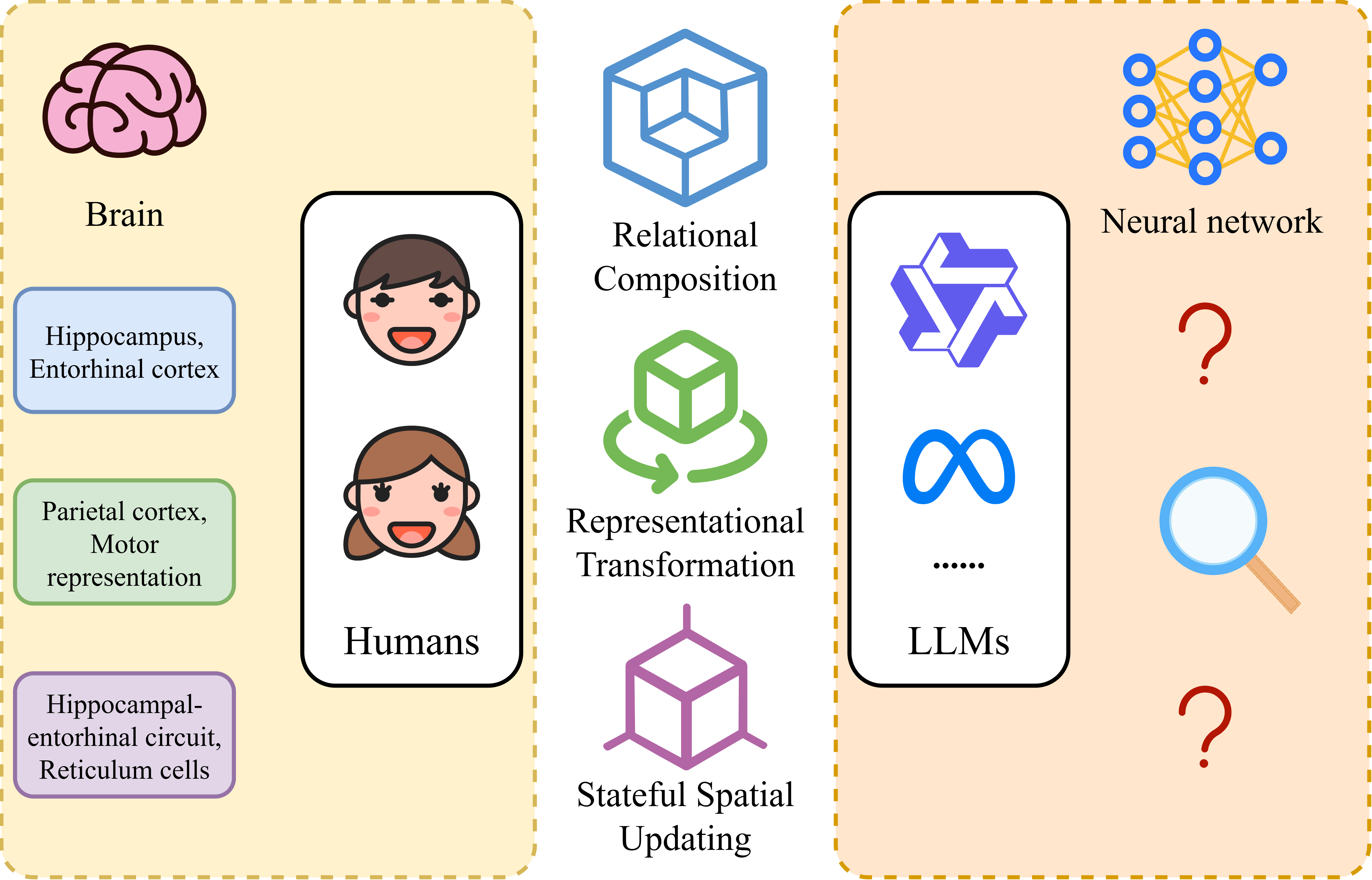}
  \caption{From human spatial cognition to spatial representations in large language models.}
  \label{fig:figure-1}
\end{figure}

Large language models (LLMs) and vision–language models (VLMs) have achieved rapid progress in reasoning, planning, and interactive decision making, and are increasingly deployed in settings that require spatial understanding, such as instruction following, navigation, embodied interaction, and robotics\cite{Wang2024LargeLM, Brohan2023RT2VM, Driess2023PaLMEAE, Guo2024LargeLM}. At the same time, the study of world models has emphasized the importance of structured internal representations for prediction and control\cite{Yi2018ModelbasedRL}. These developments raise a fundamental question: do foundation models acquire genuine spatial representations, or do they succeed through surface-level linguistic regularities?\cite{Yamada2023EvaluatingSU, Du2024EmbSpatialBenchBS, Li2024ReframingSR, Du2024EmbSpatialBenchBS}

Recent work has begun to address this question using spatial reasoning benchmarks, reporting steady performance gains with model scale and instruction tuning\cite{Yamada2023EvaluatingSU, Li2024ReframingSR, Wei2021FinetunedLM}. However, benchmark accuracy alone provides limited insight into underlying mechanisms. Correct input–output behavior may arise from linguistic heuristics, memorization, or prompt-induced reasoning strategies, making it unclear whether models rely on structured spatial representations or shallow correlations\cite{Holliday2014RightFT, Geirhos2020ShortcutLI, Xie2024OnMO, Wei2022ChainOT}.

This stands in contrast to research on human spatial cognition, where spatial ability is characterized not only by behavior but by well-studied internal mechanisms, such as cognitive maps, mental rotation, and path integration. These representations explain generalization, robustness, and systematic errors. By comparison, most studies of spatial reasoning in large neural models do not directly examine internal representations, leaving the basis of their spatial behavior largely opaque\cite{Yamada2023EvaluatingSU, Li2024ReframingSR, Hewitt2019ASP, Elazar2020AmnesicPB}.

The lack of mechanistic understanding has both practical and conceptual implications. As spatial reasoning becomes increasingly embedded in downstream systems, limited insight into internal representations hinders robustness assessment, interpretability, and principled comparison with biological cognition\cite{Yamins2016UsingGD}. Without representational evidence, behavioral similarities between models and humans remain speculative\cite{Firestone2020PerformanceVC}.

In this work, we argue that progress on spatial reasoning in foundation models requires moving beyond benchmark-centric evaluation toward mechanistic analysis. Rather than asking only whether a model produces correct answers, we investigate whether it develops structured, compositional representations that play a functional role in behavior. Drawing inspiration from cognitive science, we decompose spatial ability into a small set of core computational primitives and design controlled task families that isolate these primitives under standard inference settings, without eliciting explicit reasoning traces.

We introduce three families of spatial tasks: relational spatial reasoning, perspective transformation, and spatial program execution. To disentangle spatial representations from linguistic form, we construct parallel tasks in English, Chinese, and Arabic. We analyze both behavior and internal representations using probing, sparse autoencoder–based feature analysis, and causal interventions. This framework enables us to localize spatial information within models, characterize representational differences across task families, and assess their dependence on language. By grounding spatial evaluation in cognitive theory and mechanistic interpretability, our study clarifies the nature and limits of spatial ability in foundation models, and provides a foundation for moving beyond benchmark accuracy.

% Our core contributions are as follows:
% \begin{itemize}
%     \item We propose a cognitively motivated decomposition of spatial ability into core computational primitives and design controlled task families to isolate them.
%     \item We conduct a cross-linguistic evaluation of spatial representations using parallel tasks in English, Chinese, and Arabic.
%     \item We provide mechanistic and causal analyses of spatial representations in large language models, going beyond benchmark performance.
% \end{itemize}

%%% 2
\section{Related Work}

\subsection{Spatial Cognition and Computational Mechanisms}

Spatial cognition has long been studied in cognitive psychology and neuroscience as a core component of human intelligence. Classical psychometric frameworks distinguish abilities such as spatial perception, mental rotation, and visualization\cite{Ekstrom1976ManualFK, Shepard1971MentalRO}, while later work emphasizes that spatial behavior arises from distinct internal representations and computational mechanisms rather than a single faculty\cite{Eckardt1980TheHA, burgess2008spatial}.

Research on cognitive maps shows that humans and animals construct structured, allocentric representations that support relational inference beyond immediate perception\cite{Tolman1948CognitiveMI}. Studies of mental rotation and perspective taking demonstrate that these representations can undergo continuous geometric transformations\cite{Shepard1971MentalRO}, while navigation research highlights mechanisms for maintaining and updating spatial state over time\cite{Etienne2004PathII}. Together, these findings motivate decomposing spatial cognition into a small set of core computational primitives. Our work adopts this computational perspective to guide task design, rather than directly replicating psychometric categories.

\subsection{Spatial Reasoning in NLP and Large Language Models}

Spatial reasoning in NLP has traditionally focused on interpreting spatial relations expressed in language, such as prepositions, relative positions, and instructions. With the rise of large language models (LLMs), recent work has increasingly relied on benchmarks that evaluate spatial reasoning via output accuracy\cite{Suzgun2022ChallengingBT}.

These studies show that LLMs can achieve non-trivial performance on text-based spatial tasks, especially with chain-of-thought prompting or explicit intermediate reasoning. However, they also report limitations such as sensitivity to surface form variation, degraded performance under increased compositionality, and poor generalization\cite{Wei2022ChainOT, li2024llms}. Despite these observations, most work remains behavioral, offering limited insight into whether correct outputs reflect structured spatial representations or shallow linguistic heuristics\cite{Geirhos2020ShortcutLI}. In contrast, our work targets internal representations rather than output accuracy alone.

\subsection{Spatial Reasoning Benchmarks for Large Language Models}

A growing body of work proposes benchmarks to evaluate spatial reasoning in LLMs, focusing on spatial relations, compositional descriptions, and navigation-like inference in text\cite{Yamada2023EvaluatingSU, Rizvi2024SpaRCAS}. Synthetic and semi-natural datasets such as StepGame\cite{Shi2022StepGameAN} and SpatialEval\cite{Wang2024IsAP} are widely used to probe multi-step spatial inference, typically using accuracy-based evaluation.

Across benchmarks, recent studies report improved performance with prompting strategies, but also reveal substantial weaknesses, including sensitivity to paraphrasing, compositional complexity, and distribution shift\cite{Sharma2023ExploringAI}. Similar patterns appear in multimodal and vision-language models, where explicit reference structures or visualization-style reasoning can boost performance but are often tightly coupled to the evaluation setup\cite{Wu2024MindsEO, Liao2024ReasoningPW}. Surveys highlight a persistent gap between behavioral success and evidence for structured, compositional spatial representations\cite{Liu2025SpatialRI, Zheng2025MultimodalSR}. Overall, existing benchmarks largely emphasize output behavior, providing limited insight into underlying representations.

\subsection{Probing and Mechanistic Analysis of LLMs}

A parallel line of research investigates what information is encoded inside neural language models\cite{Maennel2020WhatDN, Krause2024ItemsOR, Yamada2023EvaluatingSU, nanda2022transformerlens, bloom2024saetrainingcodebase}. Linear probing tests whether variables can be decoded from hidden states, while more recent work analyzes representation geometry or identifies interpretable features using sparse autoencoders\cite{Hewitt2019ASP, Hewitt2019DesigningAI, Gupta2023StructuringRG, Raghu2017SVCCASV, Yan2024EncourageOI}. These methods also enable causal interventions that modify internal representations and measure behavioral effects\cite{Meng2022LocatingAE}.

While mechanistic analysis has been applied to many linguistic and semantic phenomena, its application to spatial reasoning remains limited\cite{Olsson2022IncontextLA, elhage2022toy}. Our work builds on these interpretability tools to analyze spatial representations in LLMs, aiming to move beyond correlational evidence toward functional and causal understanding.

In summary, prior work on spatial reasoning in LLMs has largely focused on benchmark-based behavioral evaluation, while mechanistic interpretability studies have rarely targeted spatial cognition. Our work bridges these lines by grounding task design in computational theories of spatial cognition and applying mechanistic analysis to probe whether LLMs implement core spatial primitives.

%%% 3
\section{Task Taxonomy: A Computational Decomposition of Spatial Ability}

To distinguish genuine spatial computation from surface-level linguistic pattern matching in large language models, we introduce a task taxonomy grounded in a computational decomposition of spatial ability. Rather than organizing tasks by domain or surface form, we classify them by the minimal computational primitives required for correct inference, drawing on established findings in human spatial cognition.

\begin{figure}[t]
  \includegraphics[width=\columnwidth]{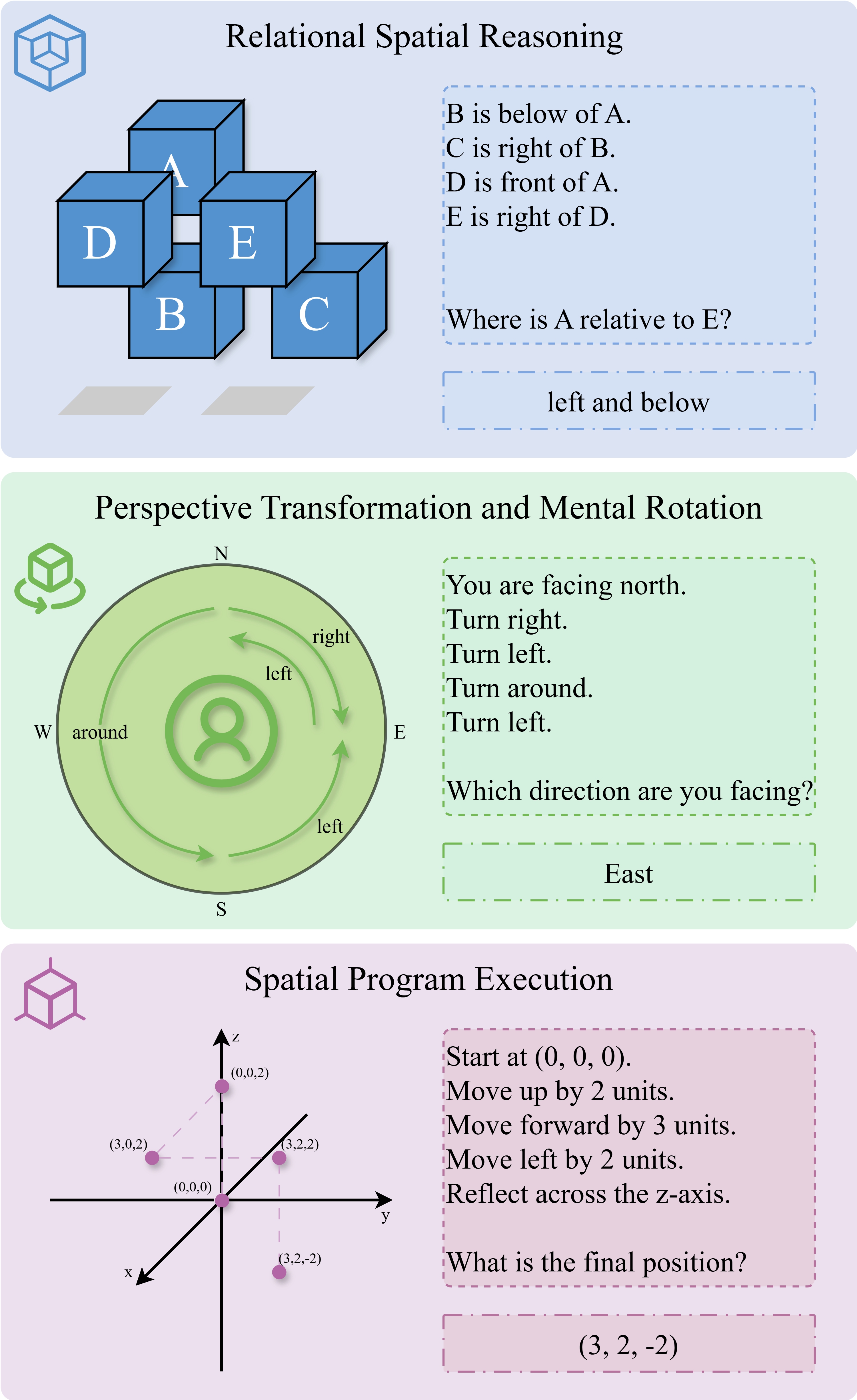}
  \caption{Illustration of the three spatial task families proposed in this work, with examples shown in English for clarity.}
  \label{fig:figure-2}
\end{figure}

We identify three irreducible components of spatial computation:
(1) relational composition,
(2) representational transformation, and
(3) stateful spatial updating.
Each component defines one task family, forming a compact and mechanistically accessible test suite for probing internal spatial representations beyond linguistic heuristics. Figure~\ref{fig:figure-2} summarizes the three families with representative examples.

\subsection{Design Principles}

All task families follow four shared principles.
\textbf{Abstraction:} tasks use abstract entities and relations, minimizing reliance on world knowledge.
\textbf{Compositionality:} solving each task requires integrating multiple spatial constraints or operations.
\textbf{Parameterizability:} difficulty is systematically controlled via factors such as entity count, steps, or dimensionality.
\textbf{Mechanistic Accessibility:} intermediate spatial variables are explicitly defined, enabling probing and causal intervention.

All tasks are evaluated under standard single-pass inference, without eliciting explicit reasoning traces, and are generated using controlled rule-based procedures to ensure computational equivalence across tasks and languages (Appendix~\ref{sec:appendix-A}).

\subsection{Relational Spatial Reasoning}

This task family targets relational composition, requiring models to construct a globally consistent spatial structure from multiple pairwise relations (e.g., \emph{A left of B}, \emph{B above C}) and infer relations between indirectly connected entities. No metric information or procedural sequence is provided; the challenge lies entirely in relational structure building. We include both 2D and 3D variants, probing whether models form structured spatial representations rather than relying on memorized linguistic templates.

\subsection{Perspective Transformation and Mental Rotation}

This family isolates representational transformation. Given an initial spatial configuration or reference frame, models must apply one or more global transformations—such as rotations, reflections, or viewpoint changes—and report the resulting orientation or relation. Unlike relational reasoning, structure is assumed and must be transformed while preserving internal consistency. Both self-centered and multi-agent perspective-taking variants are included, directly probing equivariance under geometric operations.

\subsection{Spatial Program Execution}

The third family targets stateful spatial updating. Each instance specifies an initial position and a sequence of movement or transformation commands; the model must compute the final position after executing all steps. Performance depends on maintaining and updating a latent spatial state over time, making errors cumulative and diagnostic of state-tracking failures. This family abstracts the computational core of navigation and path integration and is especially amenable to causal analysis via intervention on intermediate states.

\subsection{Alignment with Human Spatial Abilities}

Although computationally motivated, the taxonomy aligns naturally with distinctions in human spatial cognition: relational reasoning with cognitive maps and spatial visualization, perspective transformation with mental rotation, and spatial program execution with dynamic updating in navigation. This alignment serves as an interpretive reference rather than a claim of equivalence.

\subsection{Cross-Linguistic Task Construction}

To disentangle spatial computation from language-specific cues, all task families are independently constructed in English, Chinese, and Arabic. Rather than direct translation, we ensure computational equivalence while allowing natural variation in surface realization, treating language as a controlled variable for analyzing whether spatial representations are shared or language-dependent.

%%% 4
\section{Methodology and Experiments}

\begin{figure*}[t]
  \includegraphics[width=1\linewidth]{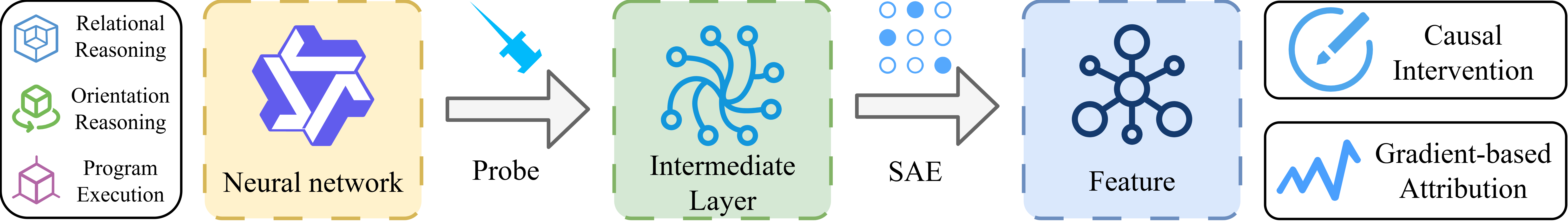}
  \caption {Overview of the proposed framework. We probe intermediate representations of a neural network, extract interpretable features using a sparse autoencoder, and analyze their roles via gradient-based attribution and causal interventions.}
  \label{fig:0}
\end{figure*}

\subsection{Experimental Setup}

\paragraph{Models.}
We evaluate spatial reasoning in two multilingual LLM families, Qwen2.5-7B\cite{Yang2024Qwen25TR} and Llama-3-8\cite{Dubey2024TheL3}, including both base and instruction-tuned variants. All experiments use publicly available checkpoints without task-specific fine-tuning, ensuring that observed spatial behaviors reflect general representations rather than adaptation.

\paragraph{Inference Protocol.}
All tasks are evaluated using standard single-pass inference without chain-of-thought prompting or explicit reasoning traces. Models select answers in a multiple-choice format by emitting a single token. This design isolates implicit spatial representations encoded in activations, rather than reasoning strategies expressed in generated text.

\paragraph{Data and Evaluation.}
For each task family and language, we construct 2,000 training instances and 200 held-out test instances, uniformly distributed across difficulty levels. Test data are never used during probing or SAE training. Unless stated otherwise, results are reported on in-distribution test sets.

\paragraph{Metrics.}
We report behavioral accuracy and analyze performance as a function of task complexity. For representational analysis, we use linear probe $R^2$ scores, regression error (MAE/RMSE for continuous variables), and layer-wise trends to characterize where spatial information emerges.

\paragraph{Multilingual Design.}
All tasks are independently constructed in English, Chinese, and Arabic. Prompts are not translated; instead, computational equivalence is preserved while allowing natural linguistic variation. Models are evaluated across all languages without language-specific tuning.

Table~\ref{tab:dataset-stats} summarizes dataset statistics.

\begin{table}[t]
\centering
\small
\setlength{\tabcolsep}{6pt}
\begin{tabular}{lccc}
\toprule
\textbf{Task Family} &
\textbf{Train} &
\textbf{Test} &
\makecell{\textbf{Difficulty} \\ \textbf{Range}} \\
\midrule
Relational Reasoning  & 2\,000 & 200 & 3--10 \\
Orientation Reasoning & 2\,000 & 200 & 2--10 \\
Program Execution    & 2\,000 & 200 & 2--10 \\
\midrule
\textbf{Total (per language)} & \textbf{6\,000} & \textbf{600} & -- \\
\textbf{Total (3 languages)}  & \textbf{18\,000} & \textbf{1\,800} & -- \\
\bottomrule
\end{tabular}
\caption{Dataset statistics across task families. Each task family is constructed in English, Chinese, and Arabic with computational equivalence.}
\label{tab:dataset-stats}
\end{table}

\begin{table*}[t]
\centering
\small
\setlength{\tabcolsep}{6pt}
\renewcommand{\arraystretch}{1.15}
\begin{adjustbox}{max width=\textwidth}
\begin{tabular}{llccc}
\toprule
\textbf{Model} & \textbf{Lang} &
\textbf{Relational} &
\textbf{Orientation} &
\textbf{Program Execution} \\
\midrule
\multirow{3}{*}{Qwen2.5-7B-Base}
& EN & 50.0 (100/200) & 23.0 (46/200) & 55.5 (111/200) \\
& ZH & 39.5 (79/200)  & 23.5 (47/200) & 58.0 (116/200) \\
& AR & 39.5 (79/200)  & 24.5 (49/200) & 47.0 (94/200) \\
\midrule
\multirow{3}{*}{Qwen2.5-7B-Instruct}
& EN & 49.0 (98/200) & 28.0 (56/200) & 62.0 (124.0/200) \\
& ZH & 47.5 (95/200) & 28.0 (56/200) & 52.5 (105/200) \\
& AR & 31.0 (62/200) & 26.0 (52/200) & 57.5 (115/200) \\
\midrule
\multirow{3}{*}{Llama3-8B-Instruct}
& EN & 4.0 (8/200)  & 2.0 (4/200)  & 2.0 (4/200) \\
& ZH & 0.0 (0/200)  & 2.5 (5/200)  & 0.0 (0/200) \\
& AR & 3.5 (7/200)  & 3.0 (6/200)  & 0.5 (1/200) \\
\bottomrule
\end{tabular}
\end{adjustbox}
\caption{Accuracy across task families and languages (\%).
Values are reported as accuracy with raw counts in parentheses.
For Program Execution, accuracy may exceed 100\% due to partial-credit scoring. Program Execution shows the strongest and most consistent performance across languages, while Orientation Reasoning performs near chance level. Arabic exhibits lower performance on Relational Reasoning, possibly reflecting language-specific encoding challenges.}
\label{tab:performance-summary}
\end{table*}

Performance does not degrade monotonically with task complexity. In relational reasoning and perspective transformation, accuracy exhibits non-monotonic fluctuations across step lengths, suggesting regime shifts in internal processing rather than simple capacity limits. In contrast, spatial program execution shows more stable degradation, indicating that coordinate-based updating is comparatively more learnable from language statistics.

Cross-linguistic evaluation reveals moderate language dependence. English and Chinese show comparable performance overall, while Arabic lags most noticeably on relational reasoning tasks. Notably, spatial program execution exhibits the smallest cross-linguistic gap, suggesting greater language invariance for coordinate-based representations. As shown in Figure~\ref{fig:0}

\begin{table*}[t]
\centering
\small
\setlength{\tabcolsep}{5pt}
\renewcommand{\arraystretch}{1.15}
\begin{adjustbox}{max width=\textwidth}
\begin{tabular}{llccc}
\toprule
\textbf{Model} & \textbf{Lang} &
\makecell{\textbf{Relational} \\ \textbf{Reasoning}} &
\makecell{\textbf{Orientation} \\ \textbf{Reasoning}} &
\makecell{\textbf{Program} \\ \textbf{Execution}} \\
\midrule

\multirow{3}{*}{Llama3-8B-Instruct}
& EN & $R^2$=.382 / $L$=25 / MAE=.55 / RMSE=.70
     & $R^2$=–.007 / $L$=0
     & $R^2$=.394 / $L$=26 / MAE=1.83 / RMSE=2.79 \\
& ZH & $R^2$=.254 / $L$=22 / MAE=.59 / RMSE=.77
     & $R^2$=–.003 / $L$=0
     & $R^2$=.347 / $L$=24 / MAE=1.93 / RMSE=2.90 \\
& AR & $R^2$=.272 / $L$=24 / MAE=.59 / RMSE=.76
     & $R^2$=–.011 / $L$=0
     & $R^2$=.347 / $L$=24 / MAE=1.93 / RMSE=2.90 \\
\midrule

\multirow{3}{*}{Qwen2.5-7B-Instruct}
& EN & $R^2$=.366 / $L$=19 / MAE=.55 / RMSE=.71
     & $R^2$=–.008 / $L$=9
     & $R^2$=.402 / $L$=20 / MAE=1.86 / RMSE=2.80 \\
& ZH & $R^2$=.243 / $L$=19 / MAE=.60 / RMSE=.78
     & $R^2$=–.006 / $L$=0
     & $R^2$=.457 / $L$=20 / MAE=1.65 / RMSE=2.67 \\
& AR & $R^2$=.264 / $L$=20 / MAE=.59 / RMSE=.76
     & $R^2$=–.005 / $L$=0
     & $R^2$=.418 / $L$=20 / MAE=1.82 / RMSE=2.74 \\
\midrule

\multirow{3}{*}{Qwen2.5-7B-Base}
& EN & $R^2$=.305 / $L$=31 / MAE=.57 / RMSE=.75
     & $R^2$=–.001 / $L$=17
     & $R^2$=.168 / $L$=31 / MAE=2.14 / RMSE=3.27 \\
& ZH & $R^2$=.207 / $L$=31 / MAE=.60 / RMSE=.80
     & $R^2$=–.003 / $L$=0
     & $R^2$=.156 / $L$=31 / MAE=2.19 / RMSE=3.27 \\
& AR & $R^2$=.172 / $L$=31 / MAE=.63 / RMSE=.82
     & $R^2$=–.002 / $L$=4
     & $R^2$=.156 / $L$=31 / MAE=2.19 / RMSE=3.27 \\

\bottomrule
\end{tabular}
\end{adjustbox}

\caption{Best-layer probing results by model, language, and task family.
We report coefficient of determination ($R^2$), best-performing layer ($L$),
mean absolute error (MAE), and root mean squared error (RMSE).}
\label{tab:probe_big_table}
\end{table*}

\subsection{Probing Spatial Representations}
We train linear probes to decode task-relevant spatial variables from model hidden states, including relative position vectors (Task~1), orientation vectors (Task~2), and absolute spatial coordinates (Task~3). Probes are trained on the training split and evaluated on held-out test instances. 

\begin{figure}[h]
  \centering
  \includegraphics[width=\columnwidth]{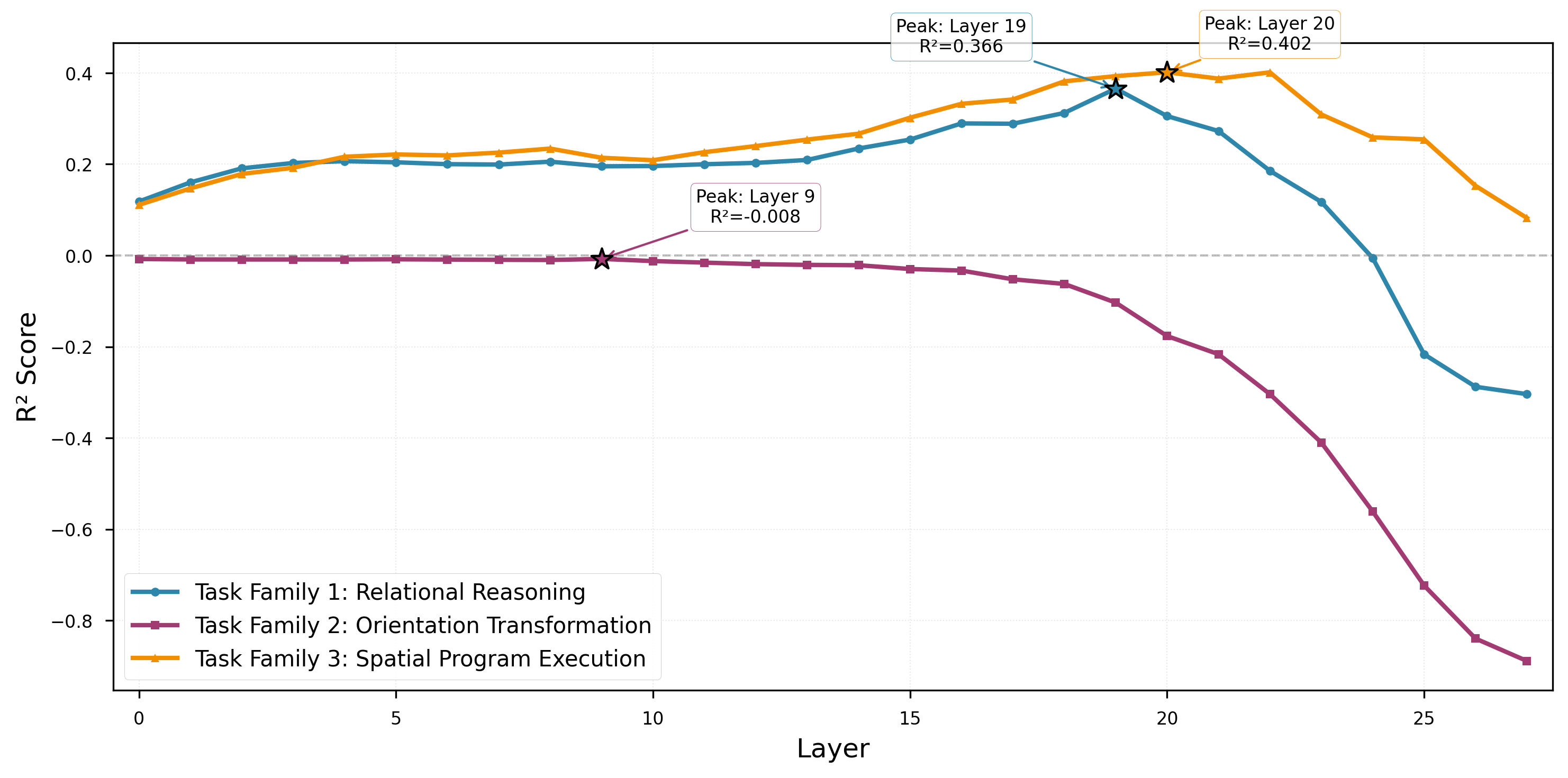}
  \caption{Layer-wise R² scores for spatial variable prediction across three task families (Qwen2.5-7B-Instruct, English). All tasks show mid-layer peaks followed by sharp declines in final layers. Task Family 1 and 3 demonstrate strong representational clarity (R² up to 0.37 and 0.40 respectively), while Task Family 2 shows minimal spatial encoding.}
  \label{fig:figure-4}
\end{figure}

We train linear probes to decode task-relevant spatial variables from hidden states, including relative position vectors (Task~1), orientation vectors (Task~2), and absolute coordinates (Task~3). Probes are trained on training instances and evaluated on held-out test data. Across models and tasks, spatial information consistently peaks in intermediate layers and declines sharply toward the final layers (Figure~\ref{fig:figure-4}). For Qwen2.5-7B-Instruct, relational reasoning reaches a maximum $R^2$ of 0.37 at layer 19, while spatial program execution peaks at $R^2 \approx 0.25$ around layer 16. Orientation variables are weakly decodable throughout ($R^2 < 0.15$). As shown in Table~\ref{tab:probe_big_table}.

This inverted-U pattern indicates that spatial representations are constructed during intermediate processing but are not preserved into the final layers responsible for token prediction. Instruction-tuned models consistently show stronger spatial representations than base models, while cross-linguistic comparisons reveal similar layer-wise emergence patterns with varying representational strength.

\subsection{Sparse Autoencoder Analysis}

\begin{figure}[h]
  \centering
  \includegraphics[width=\columnwidth]{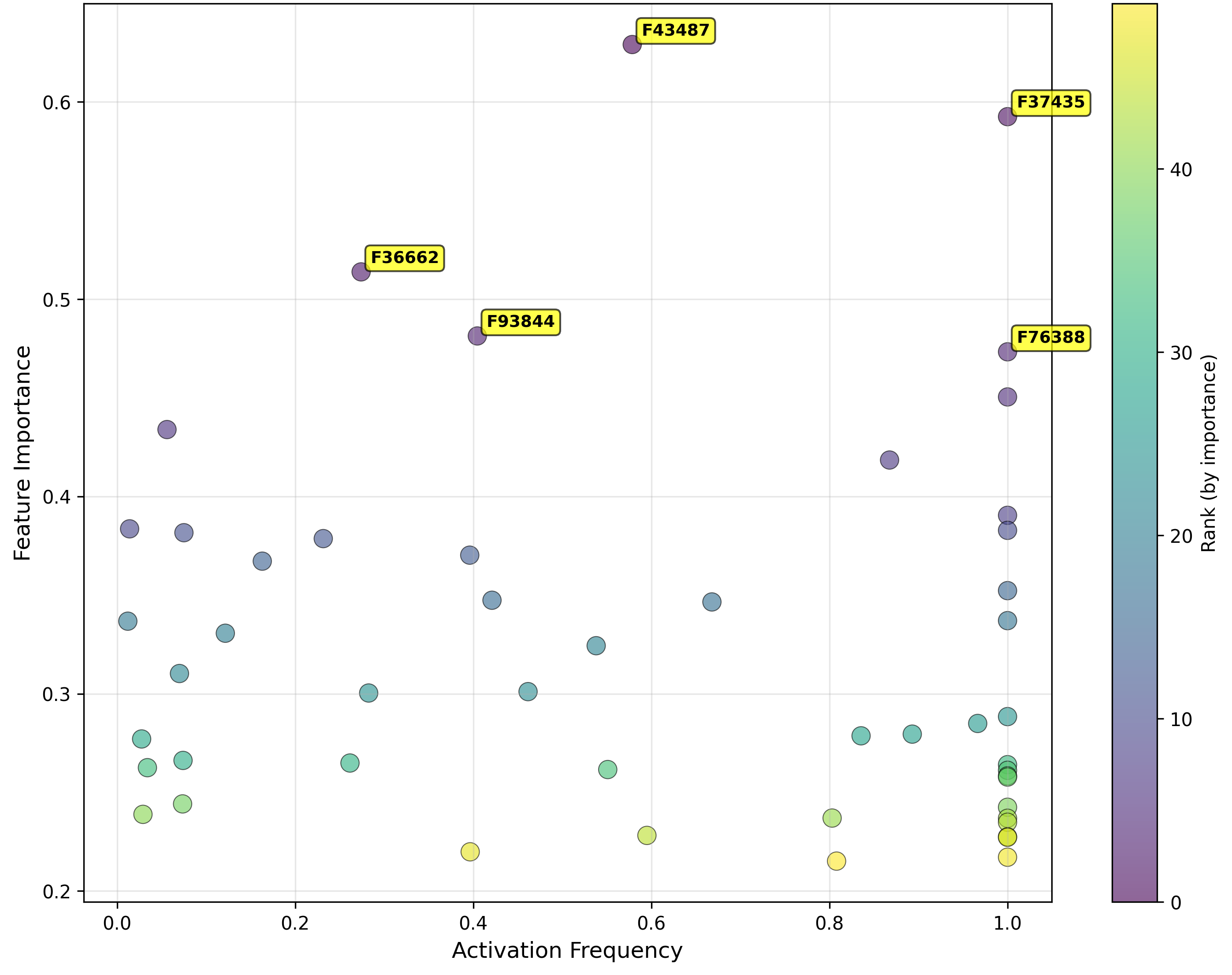}
  \caption{Feature importance versus activation frequency for Task Family 2 (Qwen2.5-7B-Istruct, Chinese). Important features are sparse and not aligned with activation frequency, indicating dissociation between usage and causal contribution.}
  \label{fig:figure-5}
\end{figure}

To identify interpretable spatial features, we train sparse autoencoders (SAEs) on layers with peak probe performance for each task family. SAEs reveal a small subset of spatially selective features, typically accounting for 3--5\% of all discovered features. As shown in Figure~\ref{fig:figure-5}.

Across tasks, we observe clear axis- or direction-selective features, including units specialized for cardinal orientations (Task~2), coordinate ranges (Task~3), and relational axes (Task~1). Gradient-based attribution shows that features with stronger spatial selectivity contribute disproportionately to spatial predictions.

Feature overlap across task families is limited (12--18\%), suggesting that different forms of spatial computation rely on largely distinct feature sets. Cross-linguistic analysis reveals partial feature sharing based on activation frequency, but substantially lower overlap when features are ranked by causal attribution, indicating language-specific mechanistic pathways supporting functionally similar behavior.

\subsection{Causal Interventions}

We perform activation patching and SAE feature ablation to assess whether spatial representations causally influence model behavior.

\begin{figure}[h]
  \centering
  \includegraphics[width=\columnwidth]{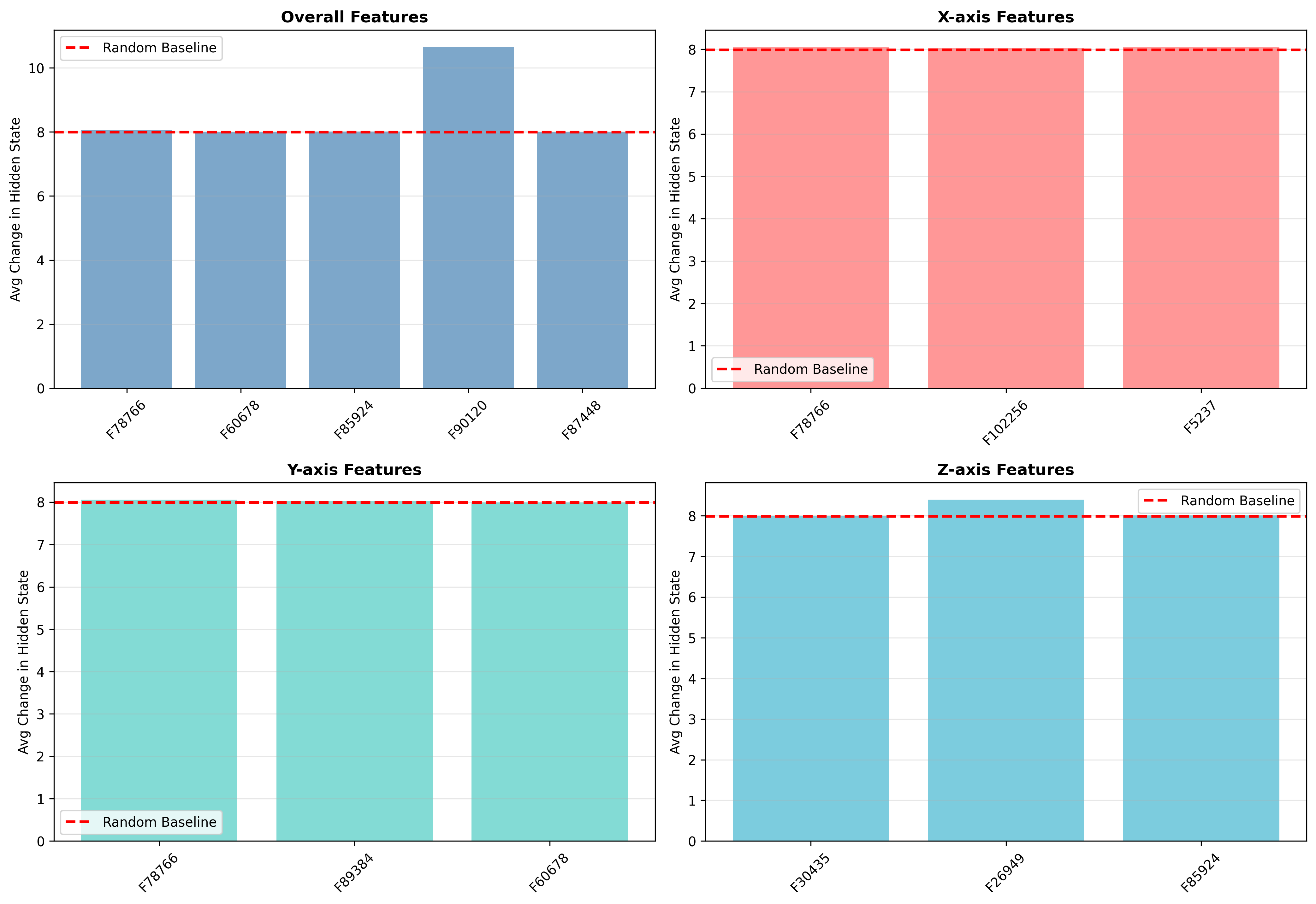}
  \caption{Intervention effects of spatial features across dimensions (Qwen2.5-7B-Instruct, Task Family 3, English). Bars show the average hidden-state change caused by intervening on spatially selective features, with the dashed line indicating a random-feature baseline.}
  \label{fig:figure-6}
\end{figure}

In spatial program execution, patching intermediate-layer activations with counterfactual spatial states systematically shifts model outputs toward counterfactual trajectories, with strongest effects in layers 14--18. Interventions at final layers show minimal impact, despite being closest to output generation. As shown in Figure~\ref{fig:figure-6}.

Complementary SAE feature ablation confirms functional specificity. Removing top spatial features reduces accuracy by 29\% in orientation tasks and 14.5\% in spatial program execution, while ablating non-spatial control features has negligible effect. These results establish that identified spatial representations are not merely decodable, but causally contribute to behavior.

%%% 5
\section{Analyses and Discussion}

\subsection{What Kind of Spatial Representations Do LLMs Develop?}

A central question of this work is whether large language models develop structured spatial representations analogous to cognitive maps, or whether spatial behavior arises from shallow linguistic heuristics. Our results suggest an intermediate regime: LLMs do encode spatial information, but these representations are conditional, fragmented, and weakly integrated into decision-making.

Linear probing reveals that task-relevant spatial variables are decodable from intermediate layers, with peak $R^2$ values up to 0.37--0.40 for relational reasoning and spatial program execution. These representations exhibit geometric coherence, supporting multi-axis integration rather than isolated symbolic associations. Crucially, causal interventions confirm functional relevance: activation patching at mid-layers reliably shifts model outputs toward counterfactual spatial trajectories.

At the same time, spatial representations are fragile and transient. Probe performance drops sharply in final layers, indicating that spatial structure constructed during intermediate processing is not preserved through output generation. Moreover, spatial features are largely task-specific, with limited overlap across relational, orientation, and coordinate-based reasoning. This fragmentation contrasts with biological cognitive maps, which support multiple spatial computations through a shared representational substrate.

Taken together, these findings indicate that LLMs exhibit \emph{conditional and partial spatial encoding}: spatial structure is constructed when task format and linguistic cues make it accessible, but is neither persistent nor unified across tasks.

\subsection{Failure Modes Across Spatial Tasks}

Distinct failure patterns across task families further illuminate the limits of LLM spatial reasoning.

\paragraph{Relational reasoning: compositional collapse.}
In relational tasks, accuracy degrades non-monotonically with chain length, with sharp drops at intermediate complexity. Errors frequently preserve one spatial axis while losing another, suggesting partial breakdowns in multi-constraint composition. Performance recovery at higher complexity likely reflects a shift toward coarse heuristics rather than robust relational inference.

\paragraph{Orientation transformation: representational absence.}
Orientation tracking performs near chance and exhibits near-zero probe $R^2$ across all layers. Error patterns reveal strong distributional bias toward frequent directions (e.g., north, east), indicating reliance on linguistic priors rather than geometric state tracking. Unlike position or coordinates, heading direction does not appear to be represented as a manipulable internal variable.

\paragraph{Spatial programs: arithmetic dominance.}
Spatial program execution shows the strongest behavioral performance, but error analysis reveals that a substantial fraction of failures are purely arithmetic. Once arithmetic errors are factored out, spatial performance aligns more closely with relational reasoning. This suggests that success in coordinate-based tasks partly reflects numerical competence rather than dedicated spatial computation.

\subsection{Language Dependence of Spatial Representations}

Our multilingual evaluation reveals a layered pattern of language dependence. Behaviorally, performance varies by up to 14 points across languages, most prominently in relational reasoning. Coordinate-based tasks show the smallest cross-linguistic gap, consistent with numerical formats being more language-invariant.

At the representational level, spatial information emerges at similar layers across languages, indicating a shared computational architecture. However, SAE analysis shows limited overlap in causally important features across languages. While different languages achieve comparable probe performance, they do so using distinct internal features, reflecting \emph{mechanistic degeneracy}: similar outputs supported by different internal pathways.

These findings challenge the assumption that spatial reasoning in LLMs is inherently language-agnostic. Linguistic encoding substantially shapes both the accessibility and implementation of spatial computation.

\subsection{Implications for Spatial Reasoning in LLMs}

\paragraph{Spatial representations are necessary but insufficient.}
The gap between representational strength and behavioral accuracy indicates that encoding spatial variables alone does not guarantee reliable reasoning. Bottlenecks arise in propagating and integrating spatial information into final decision layers.

\paragraph{Implicit vs.\ explicit reasoning.}
Evaluating models without chain-of-thought reveals latent spatial structure, but this structure is fragile. We hypothesize that explicit reasoning stabilizes spatial representations by externalizing intermediate states, acting as a compensatory scaffold rather than merely revealing hidden competence.

\paragraph{Task format matters.}
Performance differences across task families highlight the role of format. Coordinate-based tasks benefit from arithmetic pathways, while relational and orientation tasks demand abstract composition that LLMs struggle to sustain. High performance on one format does not imply general spatial understanding.

%%% 6
\section{Conclusion}
Large language models achieve strong performance on spatial benchmarks, yet it remains unclear whether this reflects genuine spatial reasoning or reliance on linguistic heuristics. Motivated by cognitive neuroscience, we adopt a mechanistic perspective and introduce a compact task taxonomy that isolates three core spatial primitives to analyze multilingual models across languages. We find that task relevant spatial information is encoded in intermediate layers and can causally influence behavior, but these representations are fragile, transient, and highly task specific. Spatial structure typically emerges in mid layers but is weakly preserved in final decisions, and different spatial tasks rely on largely disjoint internal features, providing limited evidence for unified spatial representations. Cross linguistic analysis reveals mechanistic degeneracy, where the emergence of spatial information is largely language invariant but the internal pathways supporting spatial computation vary across languages, with greater invariance observed in coordinate based tasks. Overall, these findings place current LLMs between shallow pattern matching and robust spatial cognition as characterized in neuroscience, suggesting that more human like spatial reasoning will require approaches that better preserve and integrate spatial representations across processing stages and motivating mechanistic evaluation beyond benchmark accuracy.

% \begin{figure}[h]
%   \centering
%   \includegraphics[width=\columnwidth]{images/top_features_grad_x_act.png}
%   \caption{Gradient × activation attribution for Task Family 3 (English). Bars show the mean absolute attribution of top spatial features, indicating that a small subset of features contributes disproportionately to spatial computation.}
%   \label{fig:}
% \end{figure}

%%% 局限性
\section*{Limitations}
This study analyzes spatial representations in two multilingual language model families at the 7–8B scale, and findings may differ for larger models or alternative architectures. The proposed tasks isolate three spatial primitives and do not cover more complex spatial abilities such as planning or embodied interaction. Experiments are limited to English, Chinese, and Arabic, and spatial representations may interact differently with other languages. Causal conclusions rely on activation patching and feature ablation, which provide partial causal characterization, and representational analyses focus on linearly decodable information, potentially missing nonlinear or distributed structure. Finally, tasks are abstract and text-only, limiting conclusions about real-world or multimodal spatial reasoning.

%%% 致谢
% \section*{Acknowledgments}

%%% 参考文献

\bibliography{custom}

\newpage

%%% 附录
\appendix

\section{Supplementary Materials}
\label{sec:appendix}

This appendix provides additional details on task generation, experimental configurations, and extended results that complement the main text.

\subsection{Task Generation Details}
\label{sec:appendix-A}

We provide detailed algorithms for generating each task family. All algorithms use controlled randomization with fixed seeds to ensure reproducibility and computational equivalence across languages.

\subsubsection{Algorithm 1: Relational Spatial Reasoning}

Algorithm~\ref{alg:mh-spatial} generates multi-hop spatial reasoning instances. The key design principle is to ensure that answering the query question requires composing multiple pairwise relations, with no direct relation between the source and target entities.

\textbf{Spatial Encoding:} We use a 3D coordinate system where each atomic relation corresponds to a unit vector: $\text{left/right} = (\pm 1, 0, 0)$, $\text{behind/front} = (0, \pm 1, 0)$, $\text{below/above} = (0, 0, \pm 1)$. The vector representation enables systematic computation of transitive relations.

\textbf{Reasoning Requirement:} To guarantee multi-hop reasoning, the target entity (last in sequence) cannot directly reference the source entity (first in sequence), ensuring at least two inference steps.

% <<< Task Family 1 伪代码 <<<
\begin{algorithm}[h]
\caption{Generation of Multi-hop Spatial Reasoning Instances}
\label{alg:mh-spatial}
\begin{algorithmic}[1]
\REQUIRE Number of samples $N$, hop range $[S_{\min}, S_{\max}]$
\ENSURE Dataset $\mathcal{D}$

\STATE Define atomic spatial relations $\mathcal{R}$ and their vector encodings
\STATE Initialize empty dataset $\mathcal{D}$

\FOR{$i = 1$ \TO $N$}
    \STATE Sample hop count $S \sim \textsc{Uniform}(S_{\min}, S_{\max})$
    \STATE Create entity sequence $(e_0, e_1, \dots, e_S)$
    \STATE Set position $\mathbf{p}(e_0) \leftarrow \mathbf{0}$
    \STATE Initialize fact set $\mathcal{K} \leftarrow \emptyset$

    \FOR{$j = 1$ \TO $S$}
        \STATE Sample reference entity $r_j \in \{e_0, \dots, e_{j-1}\}$
        \STATE Sample relation $r \in \mathcal{R}$
        \STATE $\mathbf{p}(e_j) \leftarrow \mathbf{p}(r_j) + \mathbf{v}(r)$
        \STATE Add fact $(e_j\ \text{is}\ r\ \text{of}\ r_j)$ to $\mathcal{K}$
    \ENDFOR

    \STATE Let source $s \leftarrow e_0$, target $t \leftarrow e_S$
    \STATE Compute relative vector $\Delta \leftarrow \mathbf{p}(t) - \mathbf{p}(s)$
    \STATE Derive gold relation $a \leftarrow \textsc{Rel}(\Delta)$
    \STATE Construct question $q$ from $\mathcal{K}$ and $(s,t)$
    \STATE Sample distractor options $D$ from $\mathcal{R} \setminus \{a\}$
    \STATE Shuffle options $O \leftarrow \{a\} \cup D$

    \STATE Add instance $(q, a, O, \Delta, S)$ to $\mathcal{D}$
\ENDFOR

\RETURN $\mathcal{D}$
\end{algorithmic}
\end{algorithm}
% >>> >>>

\subsubsection{Algorithm 2: Orientation Reasoning}

Algorithm~\ref{alg:orientation} generates orientation reasoning instances that require tracking heading direction through a sequence of turn actions.

\textbf{Direction Encoding:} Cardinal directions are encoded as angles in the standard mathematical convention: east = 0°, north = 90°, west = 180°, south = 270°. For probing analysis, directions are further encoded as unit vectors $(\cos\theta, \sin\theta)$ to enable continuous regression.

\textbf{Turn Actions:} We define three turn actions with deterministic effects: ``Turn right'' (clockwise 90°), ``Turn left'' (counterclockwise 90°), and ``Turn around'' (180°). These correspond to rotation matrices applied to the current heading vector.

\textbf{Stateful Reasoning:} Unlike Task Family 1, which involves relational composition over a static configuration, Task Family 2 requires maintaining and updating a latent state (current orientation) across sequential operations.

% <<< Task Family 2 伪代码 <<<
\begin{algorithm}[h]
\caption{Generation of Orientation Reasoning Instances}
\label{alg:orientation}
\begin{algorithmic}[1]
\REQUIRE Number of samples $N$, step range $[S_{\min}, S_{\max}]$
\ENSURE Dataset $\mathcal{D}$

\STATE Define cardinal directions $\mathcal{D} = \{\texttt{north, east, south, west}\}$
\STATE Define turn actions $\mathcal{A}$ with rotation offsets
\STATE Initialize empty dataset $\mathcal{D}$

\FOR{$i = 1$ \TO $N$}
    \STATE Sample number of steps $S \sim \textsc{Uniform}(S_{\min}, S_{\max})$
    \STATE Sample initial orientation $o_0 \sim \mathcal{D}$
    \STATE Set current orientation $o \leftarrow o_0$
    \STATE Initialize action list $\mathcal{K} \leftarrow \emptyset$

    \FOR{$j = 1$ \TO $S$}
        \STATE Sample turn action $a_j \sim \mathcal{A}$
        \STATE Update orientation $o \leftarrow \textsc{Rotate}(o, a_j)$
        \STATE Append $a_j$ to $\mathcal{K}$
    \ENDFOR

    \STATE Let final orientation $o_S \leftarrow o$
    \STATE Construct question $q$ from $(o_0, \mathcal{K})$
    \STATE Set answer $a \leftarrow o_S$
    \STATE Encode target vector $\mathbf{t} \leftarrow (\cos\theta, \sin\theta)$ for $o_S$
    \STATE Sample distractor options $O$ from $\mathcal{D} \setminus \{a\}$
    \STATE Shuffle answer and distractors into multiple-choice options

    \STATE Add instance $(q, a, O, \mathbf{t}, S)$ to dataset
\ENDFOR

\RETURN $\mathcal{D}$
\end{algorithmic}
\end{algorithm}
% >>> >>>

\subsubsection{Algorithm 3: Spatial Program Execution}

Algorithm~\ref{alg:spatial-procedure} generates spatial program execution instances that require computing the final position after applying a sequence of geometric transformations.

\textbf{Action Space:} We include five types of operations: (1) \emph{Move} -- translate along cardinal directions; (2) \emph{Reflect} -- mirror across coordinate axes; (3) \emph{Rotate} -- rotate around axes by 90°/180°/270°; (4) \emph{Scale} -- multiply all coordinates by a factor; (5) \emph{Translate} -- add offset vector.

\textbf{Coordinate System:} All operations are defined in a 3D Cartesian coordinate system starting at origin $(0,0,0)$. The coordinate system uses right-handed convention with $x$ (left/right), $y$ (backward/forward), $z$ (down/up).

\textbf{Cumulative State:} Each operation modifies the current position, and the final answer depends on the cumulative effect of all operations. Errors in intermediate steps propagate to the final result, making this task diagnostic of stateful computation and error accumulation.

\textbf{Quality Control:} To prevent extreme coordinate values that could arise from repeated scaling, we implement rejection sampling and constrain maximum coordinate magnitudes to $\pm 50$ units.

% <<< Task Family 3 伪代码 <<<
\begin{algorithm}[h]
\caption{Generation of Spatial Procedure Execution Instances}
\label{alg:spatial-procedure}
\begin{algorithmic}[1]
\REQUIRE Number of samples $N$, step range $[S_{\min}, S_{\max}]$
\ENSURE Dataset $\mathcal{D}$

\STATE Define action space $\mathcal{A}$ (move, reflect, rotate, scale, translate)
\STATE Initialize empty dataset $\mathcal{D}$

\FOR{$i = 1$ \TO $N$}
    \STATE Sample number of steps $S \sim \textsc{Uniform}(S_{\min}, S_{\max})$
    \STATE Initialize position $\mathbf{p}_0 \leftarrow (0,0,0)$
    \STATE Set current position $\mathbf{p} \leftarrow \mathbf{p}_0$
    \STATE Initialize action sequence $\mathcal{K} \leftarrow \emptyset$

    \FOR{$j = 1$ \TO $S$}
        \STATE Sample spatial action $a_j \sim \mathcal{A}$
        \STATE Update position $\mathbf{p} \leftarrow \textsc{Apply}(\mathbf{p}, a_j)$
        \STATE Append $a_j$ to $\mathcal{K}$
    \ENDFOR

    \STATE Let final position $\mathbf{p}_S \leftarrow \mathbf{p}$
    \STATE Construct question $q$ from $(\mathbf{p}_0, \mathcal{K})$
    \STATE Set answer $a \leftarrow \mathbf{p}_S$
    \STATE Sample distractor positions $O$ by perturbing $\mathbf{p}_S$
    \STATE Shuffle correct answer and distractors into multiple-choice options

    \STATE Add instance $(q, a, O, \mathbf{p}_S, S)$ to dataset
\ENDFOR

\RETURN $\mathcal{D}$
\end{algorithmic}
\end{algorithm}
% >>> >>>

\subsection{Full Probing Results}

\subsubsection{Layer-wise Probing Patterns}

Across all experiments, we observe consistent patterns in how spatial information is distributed across model layers:

\begin{itemize}
\item \textbf{Inverted-U profile:} Spatial information peaks in intermediate layers and declines sharply in final layers, indicating that spatial representations are constructed during intermediate processing but not preserved through output generation.
\item \textbf{Task-specific variations:} Different task families show peak spatial encoding at different layer positions, suggesting distinct computational stages for different types of spatial reasoning.
\item \textbf{Cross-linguistic consistency:} The layer-wise emergence pattern is similar across languages, though the magnitude of decodable information varies.
\end{itemize}

Detailed layer-by-layer probing results are presented in the main paper (Table~\ref{tab:probe_big_table}).

\subsection{SAE Hyperparameters and Visualizations}

\subsubsection{SAE Training Configuration}

All sparse autoencoders were trained using the SAE-Lens library with the following configuration:

\begin{itemize}
\item \textbf{Architecture:} Standard one-layer SAE with ReLU activation
\item \textbf{Expansion factor:} 32× (e.g., 3584 → 114,688 features for Qwen2.5-7B)
\item \textbf{Training samples:} 2,000 task instances per language
\item \textbf{Training tokens:} Approximately 300,000--420,000 tokens depending on language
\item \textbf{Batch size:} 4,096 tokens
\item \textbf{Learning rate:} $3 \times 10^{-4}$ with linear warmup (1,000 steps) and cosine decay
\item \textbf{L1 coefficient:} $\lambda = 0.001$ (selected via validation sweep over $\{0.0001, 0.0005, 0.001, 0.005\}$)
\item \textbf{Optimizer:} AdamW with $\beta_1 = 0.9$, $\beta_2 = 0.999$
\item \textbf{Training steps:} 300--400 batches until convergence
\item \textbf{Target layer:} Peak probe layer for each task family
\end{itemize}

\subsubsection{SAE Quality Metrics}

\begin{table}[h]
\centering
\small
\begin{tabular}{llcccc}
\toprule
\textbf{Task} & \textbf{Lang} & \textbf{MSE} & \textbf{$R^2$} & \textbf{Sparsity} & \textbf{L0} \\
\midrule
\multirow{3}{*}{Task 1}
& EN & \textbf{0.136} & \textbf{0.9996} & 0.0025 & 288.3 \\
& ZH & 0.798 & 0.9965 & 0.0023 & 268.2 \\
& AR & 0.252 & 0.9989 & \textbf{0.0019} & \textbf{219.5} \\
\midrule
\multirow{3}{*}{Task 2}
& EN & 0.147 & 0.9994 & 0.0026 & 301.4 \\
& ZH & 0.823 & 0.9962 & 0.0024 & 276.8 \\
& AR & 0.268 & 0.9987 & \textbf{0.0020} & \textbf{227.3} \\
\midrule
\multirow{3}{*}{Task 3}
& EN & \textbf{0.142} & \textbf{0.9995} & 0.0027 & 294.7 \\
& ZH & 0.765 & 0.9968 & 0.0025 & 271.5 \\
& AR & 0.241 & 0.9990 & \textbf{0.0021} & \textbf{223.8} \\
\bottomrule
\end{tabular}
\caption{SAE reconstruction quality and sparsity metrics for Qwen2.5-7B-Instruct. MSE: mean squared error; $R^2$: explained variance; Sparsity: fraction of active features; L0: average number of active features per sample. Best values per task family in bold.}
\label{tab:appendix-sae-quality}
\end{table}

Reconstruction MSE ranges from 0.136 (EN, Task 1) to 0.798 (CN, Task 1), with explained variance consistently above 0.996 except for Chinese Task 1 (0.9965). Average L0 (number of active features) ranges from 219 (AR, Task 1) to 301 (EN, Task 2), indicating successful sparsity enforcement.

\subsubsection{Feature Activation Distributions}

Most features activate rarely, with a heavy-tailed distribution where a small proportion of features account for the majority of activation mass. This confirms that SAEs successfully learn sparse representations.

\subsubsection{Feature Selectivity Analysis}

We compute feature selectivity using gradient-based attribution (gradient × activation). Analysis of SAE features reveals that spatially selective features are not necessarily the most frequently activated, indicating dissociation between usage frequency and causal importance for spatial reasoning.

\subsection{Prompt Templates (English)}

All tasks follow a multiple-choice format with four options. Below are representative prompt templates for each task family in English.

\subsubsection{Task Family 1: Relational Spatial Reasoning}

% English
\begin{promptbox}
\textbf{[English]}

\texttt{System:} You are a helpful assistant.

\medskip
\texttt{User:} Given the following spatial facts, answer the question by selecting one option.

\medskip
D is behind C.\\
B is left of A.\\
C is behind B.\\
E is right of D.

\medskip
Where is E relative to A?

\medskip
A. right and behind\\
B. left and front\\
C. behind\\
D. left and behind

\medskip
\texttt{Assistant:} The answer is
\end{promptbox}

\subsubsection{Task Family 2: Orientation Reasoning}

% English
\begin{promptbox}
\textbf{[English]}

\texttt{System:} You are a helpful assistant.

\medskip
\texttt{User:} Follow the instructions step by step and answer the question by selecting one option.

\medskip
You are facing north.\\
Turn right.\\
Turn left.\\
Turn around.\\
Turn right.

\medskip
Which direction are you facing now?

\medskip
A. north\\
B. east\\
C. south\\
D. west

\medskip
\texttt{Assistant:} The answer is
\end{promptbox}

\subsubsection{Task Family 3: Spatial Program Execution}

% English
\begin{promptbox}
\textbf{[English]}

\texttt{System:} You are a helpful assistant.

\medskip
\texttt{User:} Execute the spatial operations step by step and answer the question by selecting one option.

\medskip
Start at (0, 0, 0).\\
Move forward by 3 units.\\
Move right by 2 units.\\
Reflect the position across the z-axis.\\
Move up by 1 unit.

\medskip
What is the final position?

\medskip
A. (2, 3, -1)\\
B. (2, 3, 1)\\
C. (-2, 3, 1)\\
D. (2, -3, 1)

\medskip
\texttt{Assistant:} The answer is
\end{promptbox}

\subsubsection{Inference Protocol}

For all experiments:
\begin{itemize}
\item Models generate a single token continuation after the prompt
\item The most likely single-character token (A/B/C/D) is selected as the answer
\item No chain-of-thought or explicit reasoning is elicited
\item Temperature is set to 0 (greedy decoding)
\item Maximum generation length is 1 token
\end{itemize}

This protocol ensures that evaluation targets implicit spatial representations encoded in activations rather than verbalized reasoning strategies.

\subsection{Dataset Statistics and Quality Control}

\subsubsection{Data Generation Process}

All datasets were generated using rule-based procedures with controlled randomization. For each task family:

\begin{enumerate}
\item \textbf{Entity/Action Sampling:} Entities (Task 1), directions (Task 2), and operations (Task 3) are sampled uniformly at random.
\item \textbf{Constraint Verification:} Generated instances are verified to ensure they require multi-step reasoning and do not allow shortcut solutions.
\item \textbf{Answer Distribution:} Correct answers are balanced across option positions (A/B/C/D) with uniform distribution (25\% each).
\item \textbf{Difficulty Stratification:} Instances are stratified by complexity level (number of steps) with uniform distribution across difficulty range.
\end{enumerate}

\subsubsection{Cross-Linguistic Consistency}

To ensure computational equivalence across languages, we verify:
\begin{itemize}
\item \textbf{Structural isomorphism:} Same reasoning steps and answer patterns
\item \textbf{Token count variance:} Chinese prompts are typically 15--20\% shorter due to character-level encoding; Arabic prompts are 10--15\% longer
\item \textbf{Template diversity:} All languages use the same number of prompt templates
\item \textbf{Random seed alignment:} Parallel instances across languages use the same random seed, ensuring matched difficulty levels
\end{itemize}

\subsection{Additional Experimental Details}

\subsubsection{Computational Resources}

All experiments were conducted on NVIDIA A100 GPUs (40GB). Typical resource requirements:
\begin{itemize}
\item Model inference (2,000 samples): 10--15 minutes
\item Linear probe training (all layers): 30--45 minutes
\item SAE training (one layer, one task): 2--3 hours
\item Intervention experiments: 1--2 hours per configuration
\end{itemize}

Total compute: approximately 500 GPU-hours across all models, languages, and tasks.

\subsubsection{Reproducibility}

We provide:
\begin{itemize}
\item Complete data generation code with fixed random seeds
\item Exact model checkpoints and inference configurations
\item Probe training scripts with hyperparameter specifications
\item SAE training configurations and learned feature dictionaries
\item Intervention protocols and analysis notebooks
\end{itemize}

All code will be released upon publication to ensure full reproducibility.

\subsection{Intervention Experiment Details}

\subsubsection{Activation Patching Protocol}

For each intervention experiment:

\begin{enumerate}
\item \textbf{Baseline forward pass:} Run model on original input, collect activations at target layer
\item \textbf{Counterfactual forward pass:} Run model on counterfactual input (e.g., different spatial trajectory), collect activations at same layer
\item \textbf{Patching:} Replace activations at target layer with counterfactual activations
\item \textbf{Measurement:} Continue forward pass and measure output shift toward counterfactual prediction
\end{enumerate}

We report:
\begin{itemize}
\item \textbf{Shift magnitude:} KL divergence between original and patched output distributions
\item \textbf{Prediction change:} Whether patching changes the top-1 prediction
\item \textbf{Layer specificity:} Effect size as a function of intervention layer
\end{itemize}

\subsubsection{SAE Feature Ablation Protocol}

For each feature ablation experiment:

\begin{enumerate}
\item \textbf{Feature identification:} Rank SAE features by gradient-based attribution
\item \textbf{Selective ablation:} Zero out top-$k$ features for varying $k$ values
\item \textbf{Reconstruction:} Decode modified SAE activations back to hidden states
\item \textbf{Evaluation:} Measure accuracy drop on test set
\end{enumerate}

Feature ablation experiments confirm that identified spatial features causally contribute to model behavior, with accuracy systematically decreasing as more top-ranked features are removed.

\subsubsection{Counterfactual Construction}

For spatial program execution (Task 3), we construct counterfactuals by:
\begin{itemize}
\item Flipping a single operation in the sequence (e.g., ``move left'' $\rightarrow$ ``move right'')
\item Computing the resulting position divergence $\|\mathbf{p}_{\text{actual}} - \mathbf{p}_{\text{counterfactual}}\|$
\item Selecting pairs with divergence in range [3, 10] units (neither too similar nor too different)
\end{itemize}

For relational reasoning (Task 1), counterfactuals are constructed by:
\begin{itemize}
\item Flipping one spatial relation (e.g., ``A left of B'' $\rightarrow$ ``A right of B'')
\item Verifying that the modified configuration is logically consistent
\item Computing answer change (same vs. different final relation)
\end{itemize}

\end{document}